\documentclass[fleqn,10pt]{wlscirep}
\usepackage[utf8]{inputenc}
\usepackage[T1]{fontenc}

\usepackage[utf8]{inputenc}
\usepackage{tikz}
\usetikzlibrary{positioning,arrows.meta,fit,calc}

\usepackage{cite}
\usepackage{amsmath,amssymb,amsfonts}
\usepackage{algorithmic}
\usepackage{graphicx}
\usepackage{algorithm,algorithmic}
\usepackage{hyperref}
\hypersetup{hidelinks=true}
\usepackage{textcomp}
\usepackage{booktabs}
\usepackage{newunicodechar}
\newunicodechar{～}{\textasciitilde}

\usepackage[utf8]{inputenc}
\usepackage[T1]{fontenc}
\usepackage{listings}

\usepackage{longtable}
\usepackage{multirow}
\usepackage{array}

\usepackage{xcolor}

\title{Explainable Prediction from Mobile Sensing Data through LLM-guided Concept Integration}

\author[1,*]{Yuning Wang}
\author[1]{Iman Azimi}
\author[2]{Amir M. Rahmani}
\author[1]{Pasi Liljeberg}
\affil[1]{University of Turku, Department of Computing, Turku, Finland}
\affil[2]{University of California, Irvine, Department of Computer Science, California, USA}

\affil[*]{yuning.y.wang@utu.fi}

\keywords{Mobile sensing, Explainable artificial intelligence, Large language models, Concept-based learning}

\begin{abstract}

Mobile sensing enables longitudinal monitoring of behavioral and physiological patterns in everyday settings. However, accurate prediction remains challenging in small-cohort health-sensing studies, where task-specific outcome supervision is limited relative to heterogeneous sensing data. Interpretability is also important, as model outputs should reflect meaningful behavioral and physiological patterns rather than predictive scores alone. We develop a Concept-Integrated Transformer (CIT) with LLM-guided concept supervision for explainable prediction from mobile sensing data. CIT uses a pretrained large language model to generate baseline-aware concept abnormality targets with confidence weights without manual concept annotation. Across two longitudinal datasets, CIT achieves the highest F1 score on AFFECT (0.756) and ties for the highest on a PHQ-9 dataset (0.765). The learned concept scores also reveal interpretable behavioral and physiological patterns; in AFFECT, sleep quantity and quality show the clearest difference between high and low negative affect groups. These findings support LLM-guided concept integration for accurate and interpretable prediction in small-cohort mobile sensing studies.

\end{abstract}

\begin{document}

\flushbottom
\maketitle
%
%
\thispagestyle{empty}

\section{Introduction}
\label{sec:introduction}

Mobile sensing, including data collected from wearable and smartphone technologies, enables continuous observation of behavioral and physiological states in everyday settings. Such technologies allow for the collection of continuous longitudinal data, including measures of sleep, physical activity, physiological regulation, smartphone use, and self-reported states~\cite{ruiz2021great, majumder2017wearable}. 
Such data, which capture behavioral and physiological patterns over time, create opportunities to build models and predict relevant health outcomes~\cite{shen2025passive}.

In practice, however, achieving accurate prediction from mobile sensing data remains challenging, especially in real-world small-cohort studies. 
Deep learning models have been widely used for health-related prediction from longitudinal and multimodal sensing data. 
For example, Paz-Arbaizar et al. developed a Transformer-based emotion forecasting model that used passively collected mobile sensing variables, including step count, location, and sleep patterns, to predict self-reported emotional states and detect sudden changes in emotional valence~\cite{paz2025emotion}. 
Li and Zhang proposed a CNN--Transformer architecture for affective state recognition from long-term multimodal physiological data collected in everyday settings~\cite{s25030761}. 
These studies provide solid evidence that deep learning models are effective in capturing temporal dependencies and multimodal interactions in health sensing data. 
However, their performance can still be limited in real-world small-cohort studies, where reliable outcome labels are often sparse because self-report instruments or clinical assessments are costly, burdensome, or impractical to collect repeatedly~\cite{gu2025transforming}.

In addition to predictive accuracy, interpretability remains a major challenge for health-related prediction from multimodal sensing data. Input-attribution methods are commonly used to support interpretability in health-related prediction. For example, Yang et al.~\cite{YANG2024100464} used SHapley Additive exPlanations (SHAP) to explain personalized affect-forecasting models by quantifying how wearable sensor features and self-reported diary variables contributed to individual predictions. 
Concept-based approaches provide a higher level of interpretation by relating predictions to clinically or behaviorally meaningful domains. Wu et al., for example, used a concept bottleneck framework for clinical time-series prediction, allowing model outputs to be interpreted through concepts including kidney function, respiratory and oxygenation status, and general illness severity~\cite{pmlr-v182-wu22a}.
However, input-level explanations can still be difficult to translate into participant-specific summaries in longitudinal multimodal sensing data, where the meaning of a feature may depend on the participant's baseline, temporal trend, and co-occurring behavioral or physiological signals~\cite{mohr2017personal,stamatis2024differential}. 
Concept-based models offer a more structured alternative, but their training requires concept-level supervision for individual samples or temporal windows. Although relevant concept categories can often be defined from domain knowledge, reliable concept labels are rarely available in real-world sensing datasets, and manual annotation of each longitudinal window by domain experts is costly and difficult to scale.

Large language models (LLMs) have recently been explored as external knowledge sources for model supervision. Previous studies have used LLMs to generate task labels, pseudo-labels, and textual annotations for downstream learning~\cite{tan2024large,pangakis2024knowledge}. These capabilities motivate their use as a source of weak concept supervision rather than as direct predictors of health outcomes. In longitudinal mobile sensing, we believe that LLM-derived concept supervision can help address the two challenges discussed above: improving predictive accuracy when outcome labels are sparse and enabling concept-level interpretation.


In this paper, we develop a Concept-Integrated Transformer (CIT) framework with LLM guidance for explainable health-related prediction from mobile sensing data. 
CIT uses a pretrained LLM to provide concept abnormality supervision without requiring manual concept annotations. 
Specifically, the LLM assesses baseline-aware statistical summaries and produces concept-level abnormality targets with confidence weights. These targets provide auxiliary supervision for the prediction model while defining an interpretable intermediate concept representation. The Transformer therefore learns to predict both health-related outcomes and concept scores directly from sensing data, allowing the same concept layer to support predictive learning and concept-level interpretation. The LLM is used only to construct concept supervision during model development and is not required at inference.

The main contributions of this study are as follows.
\begin{enumerate}
    \item Proposing CIT, a concept-integrated Transformer framework for explainable health-related prediction from longitudinal multimodal sensing data.

    \item Developing an annotation-free LLM-guided concept abnormality supervision procedure, in which an LLM evaluates baseline-aware statistical summaries and produces concept abnormality targets with consistency-based confidence weights, without requiring manual concept annotations.

    \item Evaluating CIT using two longitudinal multimodal sensing datasets, covering affect prediction on the AFFECT dataset and depression-related prediction on a PHQ-9 dataset.
    
    \item Indicating how the learned concept scores support class-level, population-level, and individual-level interpretation of model behavior.
\end{enumerate}

\section{Methodology}

\begin{figure*}[ht]
    \centering
    \includegraphics[width=\textwidth]{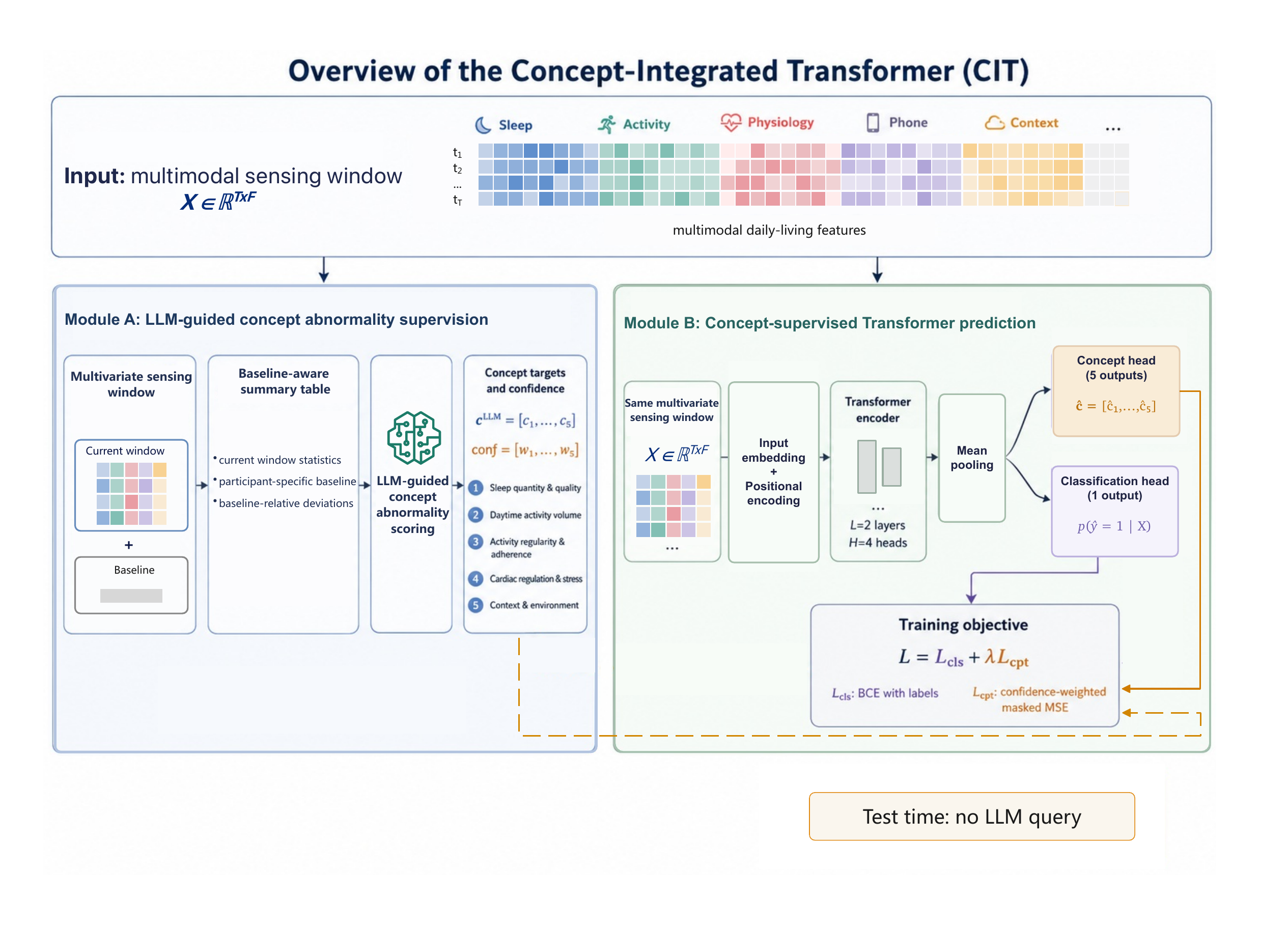}
    \caption{Overview of the Concept-Integrated Transformer (CIT). The framework uses a longitudinal multivariate sensing window $X \in \mathbb{R}^{T \times F}$, where $T$ is the window length and $F$ is the number of sensing features. The LLM-guided concept abnormality supervision module derives annotation-free concept targets and confidence weights from baseline-aware summary statistics. The concept-supervised Transformer prediction module encodes the original sensing window and jointly learns target prediction and concept prediction using a classification loss and a confidence-weighted concept loss. At test time, CIT does not query the LLM and outputs both the predicted probability and concept abnormality scores from sensing data alone.}
    \label{fig:arch}
\end{figure*}

We propose a Concept-Integrated Transformer (CIT) framework for explainable health-related prediction from longitudinal multimodal sensing data. The framework leverages concept-level supervision to guide learning under limited outcome labels, while organizing model explanations through health-relevant concepts. As shown in Figure~\ref{fig:arch}, the proposed framework consists of two connected modules: \textbf{Module A: LLM-guided concept abnormality supervision} and \textbf{Module B: concept-supervised Transformer prediction}.

The input of CIT is a multimodal sensing window constructed from longitudinal wearable and mobile sensing data. 
Each window contains observations of behavioral, physiological, and contextual features over a fixed time period and is associated with a health-related outcome label during training. 
This input window is used differently by the two modules: Module A converts it into concept supervision targets, whereas Module B uses the original time-series window for prediction learning.

Module A generates concept abnormality supervision from the multimodal sensing window. 
The sensor data is first converted into a baseline-aware statistical summary, including current window statistics, participant-specific baseline statistics, and baseline-relative deviation indicators. An LLM then assigns abnormality scores to five behavioral and physiological concepts and produces consistency-based confidence weights. These LLM-derived concept abnormality scores are not used as final predictions; instead, they serve as concept supervision signals for model training.

Module B estimates health-related outcomes with concept-level supervision.
The original multimodal sensing window is embedded, combined with positional embeddings, and encoded by a Transformer encoder. The encoded representation is passed to two prediction heads: one for the health-related label and the other one for the concept abnormality scores. During training, the classification head is supervised by the ground-truth label, while the concept head is supervised by the LLM-derived concept targets through a confidence-weighted concept loss. At test time, the LLM-guided module is no longer required. The trained Transformer directly outputs both the health prediction and the concept scores, which are used as concept-level explanations.

In the following, we describe the two modules in detail.

\subsection{Module A: LLM-guided Concept Abnormality Supervision}

This module extracts concept-level supervision signals without requiring manual concept annotations. 
For each training window, the module receives the original multivariate sensing segment and the participant's historical baseline as input, constructs a baseline-aware summary table, and assigns abnormality scores to predefined behavioral and physiological concepts, using an LLM. 
The module outputs two quantities for each sample-concept pair: an LLM-derived concept abnormality target and a consistency-based confidence weight. 
These outputs are used only during training to supervise the concept head of CIT. It should be noted that they are not treated as target labels and are not required at test time.

\subsubsection{Baseline-aware Summary Table Construction}
\label{summary_table}
Each sensing window is converted into a structured summary table that describes the recent window pattern and its deviation from the participant's own historical baseline. 
This summary table serves as the input evidence for LLM-based concept abnormality scoring.

Let $X_i \in \mathbb{R}^{T \times F}$ denote the multivariate sensing window for sample $i$, where $T$ is the window length and $F$ is the number of sensing features. 
In our experiments, $T=7$, and each window is aligned to a target day with a binary outcome label $y_i \in \{0,1\}$, where $1$ denotes the negative class. 
For each feature $f$, we compute three groups of statistics.

First, window statistics summarize the recent pattern within the current window:
\[
\mathrm{mean7}_f,\quad \mathrm{std7}_f,\quad \mathrm{slope7}_f,\quad \mathrm{last}_f,\quad \mathrm{miss7}_f .
\]
Here, $\mathrm{slope7}_f$ denotes the linear trend, $\mathrm{last}_f$ denotes the last observed value, and $\mathrm{miss7}_f$ denotes the missing rate within the window.

Second, baseline statistics describe the participant-specific historical reference pattern. 
In our experiments, we use the previous 28 days excluding the current window to compute:
\[
\mathrm{baseMean}_f,\quad \mathrm{baseStd}_f .
\]

Third, baseline-relative deviations quantify how the current window differs from the participant's baseline:
\begin{align}
    z\mathrm{Base}_f &= \frac{\mathrm{last}_f - \mathrm{baseMean}_f}{\mathrm{baseStd}_f}, \\
    z\mathrm{BaseMean7}_f &= \frac{\mathrm{mean7}_f - \mathrm{baseMean}_f}{\mathrm{baseStd}_f}, \\
    \mathrm{pctBase}_f &= \frac{\mathrm{mean7}_f - \mathrm{baseMean}_f}{|\mathrm{baseMean}_f|}, \\
    \mathrm{pctBaseLast}_f &= \frac{\mathrm{last}_f - \mathrm{baseMean}_f}{|\mathrm{baseMean}_f|}.
\end{align}

The resulting per-feature statistics are organized as key-value entries in a structured summary table. 
This table is then used as the input evidence for the LLM-guided concept abnormality scoring step described below.

\subsubsection{Generating Concept Targets and Confidence Weights}
The baseline-aware summary table is fed to the LLM to generate the two outputs: concept abnormality targets and their corresponding confidence weights. 
The concept targets provide annotation-free supervision for the concept head, while the confidence weights determine how strongly each target contributes to the concept loss.

\paragraph{LLM-guided concept abnormality scoring.}
We define five high-level concepts that summarize major behavioral and physiological dimensions in wearable and mobile sensing data:
\begin{enumerate}
    \item Sleep quantity and quality
    \item Daytime activity volume
    \item Activity regularity and adherence
    \item Cardiac regulation and stress
    \item Context and environment
\end{enumerate}
The full feature-to-concept mapping is provided in Supplementary Tables~S1 and S2.

For each training sample, we prompt an LLM, GPT-4.1-mini~\cite{openai2025gpt41}, with the summary table described above. 
The prompt is label-free: it does not include the target label or ask the LLM to infer the outcome. 
Instead, the LLM is asked to assess whether each concept appears abnormal relative to the participant's baseline evidence and to return a JSON object containing a concept abnormality score:
\begin{equation}
    c^{\mathrm{LLM}} = [c_1, \dots, c_K], 
    \quad c_k \in [0,1] \cup \{\texttt{null}\},
\end{equation}
where $K=5$ and $c_k$ represents the abnormality of concept $k$ within the current window relative to the participant's typical baseline pattern. 
A value closer to 1 indicates stronger abnormality, whereas \texttt{null} indicates that the available evidence is insufficient or ambiguous.

To keep the LLM output grounded in the provided data and reduce hallucinated evidence, we impose an evidence constraint. 
For each non-null concept score, the LLM is required to select a small set of evidence keys from the provided summary table to support its judgment, such as the 7-day linear trend in total sleep time.
If the evidence is insufficient or no valid evidence keys are provided, the corresponding concept abnormality score is treated as \texttt{null}. 
The original prompt is provided in the Supplementary Information.

\paragraph{Consistency-based Confidence Weighting}

LLM-derived concept targets can vary across repeated generations. 
Rather than relying on self-reported LLM confidence, we estimate a consistency-based confidence weight from agreement across multiple LLM runs.

For each training sample and each concept, this procedure produces two quantities: an aggregated concept abnormality target $\tilde{c}_k$ and a confidence weight $\mathrm{conf}_k$.
The aggregated target is used as the concept supervision signal, while the confidence weight is later used in the confidence-weighted concept loss so that more stable concept targets contribute more strongly to training.

For each training sample, we query the LLM $M$ times; in our experiments, $M=3$.
For each concept $k$ and repeated run $m$, the LLM returns an abnormality score $\hat{c}^{(m)}_k$ and an evidence-key set $E^{(m)}_k$.
The abnormality scores are used to estimate score consistency, whereas the evidence-key sets are used to estimate evidence consistency.
The final confidence weight combines these two consistency measures.

\textit{Abnormality score consistency.}
Let $V_k$ denote the set of non-null abnormality scores returned for concept $k$ across the $M$ runs.
When at least two valid scores are available, i.e., $|V_k|\ge 2$, we use the median score as the aggregated concept target:
\begin{equation}
\tilde{c}_k = \mathrm{median}(V_k).
\end{equation}
We then measure the dispersion of the repeated scores using the mean absolute deviation from the median:
\begin{equation}
\mathrm{MADmean}k = \frac{1}{|V_k|} \sum{v \in V_k} |v - \tilde{c}_k|.
\end{equation}
The abnormality score consistency is defined by mapping this dispersion to the interval $[0,1]$:
\begin{align}
C^{\mathrm{score}}_k &= \mathrm{clip}\!\left(1 - \frac{\mathrm{MADmean}_k}{\tau},\, 0,\, 1\right),
\end{align}
where $\tau$ is a scale hyperparameter controlling how strongly score variation reduces confidence.
In our experiments, we set $\tau=0.15$.
If fewer than two valid scores are available, we set $\tilde{c}_k=\texttt{null}$ and assign a small floor value to $C^{\mathrm{score}}_k$.

\textit{Evidence consistency.}
Let $E^{(m)}_k$ denote the evidence-key set returned for concept $k$ in run $m$.
We measure evidence consistency using pairwise Jaccard overlaps between the evidence-key sets:
\begin{equation}
J^{(i,j)}_k = \mathrm{Jaccard}\left(E^{(i)}_k, E^{(j)}_k\right).
\end{equation}
For $M=3$, this gives three pairwise overlaps.
To reduce sensitivity to a single noisy run, we define the evidence-consistency term as the mean of the two largest pairwise overlaps:
\begin{equation}
C^{\mathrm{evi}}_k = \mathrm{mean}\left(\mathrm{top2}\left\{J^{(i,j)}_k: 1 \leq i < j \leq M \right\}\right).
\end{equation}
\textit{Final confidence weight.}
The final confidence weight combines score consistency and evidence consistency:
\begin{equation}
\mathrm{conf}_k = C^{\mathrm{score}}_k \cdot \left(\alpha_k + (1-\alpha_k) C^{\mathrm{evi}}_k\right),
\label{equ}
\end{equation}
where $\alpha_k \in [0,1]$ is a soft-floor parameter that prevents evidence inconsistency from overly suppressing otherwise stable abnormality scores.

The final output of Module A for each training sample is therefore a pair 
$(c^{\mathrm{LLM}}, \mathrm{conf})$, where 
$c^{\mathrm{LLM}} \in (\mathbb{R}\cup\{\mathrm{NaN}\})^{K}$ contains the LLM-derived concept targets and 
$\mathrm{conf}\in[0,1]^K$ contains the corresponding confidence weights. 
These outputs are passed to the concept-supervised Transformer prediction module as supervision for the concept head during training.

\subsection{Module B: Concept-supervised Transformer Prediction}
\label{sec:stage2_cit}

This module develops a predictive model, receiving the original multivariate sensing window as input while aligning part of its representation with the LLM-derived concepts from Module A. 

Given a batch of sensing windows $X\in\mathbb{R}^{B\times T\times F}$, Module B outputs two quantities: 
(i) a target logit $\hat{y}_{\mathrm{logit}}\in\mathbb{R}^{B}$ for the binary health-related outcome, and 
(ii) concept abnormality predictions $\hat{\mathbf{c}}\in[0,1]^{B\times K}$, where $K=5$. 
During training, the classification head is trained to predict the ground-truth label, and the concept head is trained to predict the LLM-derived targets from Module A. The detailed network architecture is described below, and the training setup is provided in Section~\ref{sec:exp_setup}.

\subsubsection{Architecture}

As mentioned previously, Module B receives the original multivariate sensing window as input and produces two outputs: 
a target logit for the binary health-related prediction task and a vector of concept abnormality scores.

An encoder-only Transformer backbone is chosen to encode the temporal sequence of daily feature vectors. 
The encoded sequence is aggregated into a fixed-length window representation, which is then passed to two separate MLP heads. 
The classification head outputs the target logit 
$\hat{y}_{\mathrm{logit}}$, while the concept head outputs $K$ concept logits followed by a sigmoid activation to obtain 
$\hat{\mathbf{c}}\in[0,1]^K$. 
The architecture, therefore, allows the model to jointly learn the target prediction task and the concept prediction task from the same temporal representation.

\subsubsection{Joint Training Objective}
The training objective combines classification and concept supervision. 
The classification loss trains the classification head to predict the target label. 
The concept loss trains the concept head to approximate the LLM-derived concept abnormality targets, while using the confidence weights and missing-value mask to avoid over-penalizing uncertain or unavailable concept targets.

Let
$c^{\mathrm{LLM}}\in(\mathbb{R}\cup\{\mathrm{NaN}\})^{B\times K}$ denote the LLM-derived concept targets, and let $\mathrm{conf}\in[0,1]^{B\times K}$ denote the corresponding consistency-based confidence weights.
The training objective is computed by combining a classification loss with a confidence-weighted masked concept loss.

\textit{Classification loss.}
A binary cross-entropy with logits is used and defined as:
\begin{equation}
\label{equ:affect_loss}
\mathcal{L}_{\mathrm{cls}}=\frac{1}{B}\sum_{i=1}^{B}\mathrm{BCEWithLogits}
\!\left(\hat{y}^{(i)}_{\mathrm{logit}},y^{(i)}\right).
\end{equation}

\textit{Confidence-weighted masked concept loss.}
A masked mean squared error (MSE) loss is defined only over the available entries, since concept targets do not exist for every sample-concept pair. In addition, each supervised entry is weighted by its consistency-based confidence score, so that more reliable LLM-derived concept targets contribute more strongly to training.

Let $m^{(i,k)}=\mathbb{I}\!\left(c^{\mathrm{LLM}}_{i,k}\neq \mathrm{NaN}\right)$ denote the supervision mask. For implementation convenience, missing entries (i.e., NaN) in $c^{\mathrm{LLM}}$ are replaced by zeros in $\tilde{c}^{\mathrm{LLM}}$ before applying the mask. The concept loss is defined as: 

\begin{equation}
\label{equ:concept_loss}
\mathcal{L}_{\mathrm{cpt}}=
\frac{
\sum_{i=1}^{B}\sum_{k=1}^{K}
\mathrm{conf}_{i,k}m_{i,k}
\left(\hat{c}_{i,k}-\tilde{c}^{\mathrm{LLM}}_{i,k}\right)^2
}{
\sum_{i=1}^{B}\sum_{k=1}^{K}
\mathrm{conf}_{i,k}m_{i,k}+\epsilon
},
\end{equation}
where $\epsilon$ is a small constant for numerical stability. If a mini-batch contains no valid concept supervision (i.e., $\sum \mathrm{conf}\cdot m$ is near zero), we set $\mathcal{L}_{\mathrm{cpt}}=0$.

The final joint training objective is:
\begin{equation}
\label{equ:final_loss}
\mathcal{L}=\mathcal{L}_{\mathrm{cls}}+\lambda\,\mathcal{L}_{\mathrm{cpt}},
\end{equation}
where $\lambda$ is a weighting coefficient controlling the relative contribution of the concept supervision loss.

\subsubsection{Test-time Prediction and Concept Explanations}

At test time, CIT is used without any LLM query. 
Given an input window $X$, the trained Transformer outputs a target logit $\hat{y}_{\mathrm{logit}}$ and concept abnormality predictions $\hat{\mathbf{c}}\in[0,1]^K$. 
The predicted probability is computed as
\[
p(\hat{y}=1\mid X)=\sigma(\hat{y}_{\mathrm{logit}}).
\]
The concept predictions $\hat{\mathbf{c}}$ are used as concept-level abnormality indicators associated with the prediction. 
They provide a structured interpretation of the model output in terms of the five predefined health-relevant concepts used during training.
Thus, CIT retains concept-level interpretability at test time without requiring additional LLM prompting.

\section{Experimental Setup}
\label{sec:exp_setup}
\subsection{Datasets}
\label{sec:dataset}
We evaluate the proposed framework using two longitudinal multimodal sensing datasets: 1) an AFFECT dataset for negative affect prediction~\cite{labbaf2024affect_dib}, and 2) a PHQ-9 depression dataset~\cite{mohseni2026solo} to examine the framework under a different health-related target and sensing feature space.

\subsubsection{AFFECT dataset}
The AFFECT study was conducted to measure the relationships between affect and wearable data in young adults. The dataset provides longitudinal
physiological, behavioral, and mobile sensing data together with daily affect assessments collected during
the 2020 COVID-19 period \cite{labbaf2024affect_dib}.

\paragraph{Participants and study design}
The dataset included 21 undergraduate students (ages 18--22) and followed them longitudinally in 2020. Eligibility criteria included English fluency and owning an Android smartphone
(Android OS 6.0 or higher).

Participants wore an Oura ring and a Samsung Gear Sport smartwatch to capture daily activity, sleep-related,
and physiological patterns. In addition, a smartphone app (Personicle) logged behavioral context such as
activity types and location changes, with lifelog events summarized in five-minute intervals. Data collection
was integrated into the Centralive mHealth platform~\cite{centralive2026}. 

\paragraph{Negative affect label from PANAS}
Daily affect was assessed using the Positive and Negative Affect Schedule (PANAS)~\cite{watson1988panas}.
Positive Affect (PA) and Negative Affect (NA) were computed as the average
of 10 positive and 10 negative PANAS items, respectively. For binary classification, affect values were binarized relative to the participant-wise median. To reduce ambiguity among samples near the participant-specific decision boundary, we excluded the central 20\% of values, following prior work on the same dataset~\cite{jafarlou2023objective}. Note that for our experiments in this paper, we only use NA as the sole target label.

\subsubsection{PHQ-9 depression dataset}

The PHQ-9 depression dataset is from a longitudinal multimodal sensing study previously used for loneliness monitoring and depressive symptom detection~\cite{jafarlou2024objective, borelli2025detection, mohseni2026solo}. In our experiments, this dataset complements the AFFECT dataset by introducing a different health-related target, weekly depressive-symptom severity, while retaining a comparable multimodal sensing structure based on wearable, smartphone, sleep, activity, and physiological features. Given its limited sample size, we use this dataset as an experiment to assess applicability in a small-cohort setting.

\paragraph{Participants and study design}
The study recruited undergraduate students aged 18--22 in the United States. The prior depression detection study reported 28 participants after withdrawals. Participants were provided with an Oura ring and a Samsung smartwatch and installed the AWARE smartphone app, together with a survey app for daily and weekly assessments~\cite{borelli2025detection}. The sensing protocol captured daily sleep, physical activity, physiology, smartphone interaction, and mobility features, consistent with the multi-device setup described in earlier work from the same broader study~\cite{jafarlou2024objective}.

\paragraph{Depression label from PHQ-9}
Depressive symptoms were assessed weekly using a PHQ-9 questionnaire \cite{kroenke2001phq} adapted to the weekly context~\cite{borelli2025detection}. Following the prior depression detection study, PHQ-9 scores of 0--4 were assigned to the "none-minimal" group, whereas scores greater than 4 were assigned to the "follow-up-needed" group. We use this binary grouping to define the target label, where class~0 corresponds to "None-minimal" and class~1 corresponds to "follow-up-needed". For each weekly PHQ-9 assessment, the model input consists of the daily features from the preceding 7-day period.

\subsection{Data Preparation}

For the AFFECT dataset, we construct multivariate time-series samples using sliding windows. For each participant, daily features are sorted chronologically. Each sample consists of a 7-day window $\mathbf{X}_i \in \mathbb{R}^{7 \times 43}$, and the binary label $y_i$ is assigned from the observation on the window end day. For Module~A concept prompting, we additionally define a participant-specific 28-day historical baseline preceding each 7-day window, excluding the current window to avoid future information leakage.

We use a participant-level chronological 60/20/20 split for training, validation, and test based on the end day of each window. This ensures that, for each participant, validation and test samples always occur later in time than training samples. The final AFFECT dataset contains 930 training samples, 391 validation samples, and 395 test samples.

For the PHQ-9 depression dataset, each sample is defined by a weekly PHQ-9 assessment and the corresponding 7-day daily feature window preceding that assessment. The input has the form $\mathbf{X}_i \in \mathbb{R}^{7 \times 103}$. The rows in each window correspond to the daily observations from 7 days before the PHQ-9 assessment. We use the binary PHQ-9-derived target described above as $y_i$.

Given the limited sample size of the PHQ-9 experiment, we use a participant-wise chronological split in which the first 80\% of samples from each participant are used for model development and the remaining 20\% are used as the test set. Within the model-development set, we use participant-wise chronological forward validation to select the training epoch: validation blocks are always later in time than the corresponding training blocks\cite{tashman2000out}. The final model is then trained on all model-development samples and evaluated on the chronologically later test set. This results in 253 model-development samples and 77 test samples.

\subsection{Baseline Methods}

We compare the proposed CIT with three baseline methods.
The first baseline is a Transformer without a concept-learning component, which serves as a standard deep-learning-based temporal prediction model for longitudinal multimodal sensing data~\cite{paz2025emotion}. It uses the same encoder backbone as CIT, but the model contains only the classification head and does not receive any concept-level supervision. 
This comparison evaluates whether adding LLM-derived concept supervision improves prediction beyond a standard Transformer trained directly on the outcome labels.

The second baseline is a Transformer with an unsupervised concept head, which serves as an architectural ablation to test whether adding a concept branch alone improves performance. This baseline keeps the same dual-head architecture as CIT, including both the classification head and the concept head, but sets the concept-loss weight to $\lambda=0$. 
As a result, the concept head is present in the architecture, but it does not contribute to the optimization objective. 
This baseline tests whether any improvement comes from the architectural addition of a concept branch alone, rather than from the proposed confidence-weighted concept supervision.

The third baseline is CIT with OCSVM-derived concept targets, which serves as an annotation-free statistical alternative to the proposed LLM-derived concept supervision. This baseline uses the same CIT training objective but replaces the LLM-derived concept targets with concept-level anomaly scores generated by one-class support vector machines (OCSVMs)~\cite{scholkopf2001estimating}. OCSVM is a classical annotation-free method for estimating whether samples deviate from the distribution of normal data. Here, it is used as an unsupervised statistical alternative for producing concept abnormality targets. 
The comparison with the proposed framework indicates whether the proposed LLM-guided concept supervision provides benefits beyond a classical unsupervised anomaly-detection alternative.

For each predefined concept, we first select the corresponding concept-specific summary features from the baseline-aware summary table. 
An OCSVM was fitted using the training windows after median imputation and standardization. 
The OCSVM decision function was then applied to the corresponding windows, where larger decision values indicate more normal samples. 
We therefore used the negative decision value as the anomaly score. 
To obtain concept targets on the same scale as the LLM-derived scores, the anomaly scores were clipped and rescaled to $[0,1]$ using the 1st and 99th percentiles. 
Samples with fewer than 70\% non-missing concept-specific summary features were treated as unavailable for that concept. 
For available OCSVM-derived targets, the confidence weight was set to 1.

\subsection{Model Training}

All Transformer-based models (i.e., the proposed framework and the three baseline methods) use the same encoder-only backbone described in Section~\ref{sec:stage2_cit} and the same input windows as CIT to ensure a fair architectural comparison. Models are trained using Adam with a learning rate of $1\times10^{-4}$ and a weight decay of $1\times10^{-5}$. The batch size is 32 for training and 64 for validation or testing.

For the AFFECT experiment, models are trained for 300 epochs, and the checkpoint with the best validation F1 score is selected for the final testing. For the PHQ-9 experiment, the same optimization settings are used. Within the model-development set, we use participant-wise chronological forward validation to select the training epoch. The final PHQ-9 model is then trained on all model-development samples for the selected number of epochs and evaluated on the test set.

The models are optimized with the binary classification loss defined in Equation~(\ref{equ:affect_loss}). CIT additionally uses the confidence-weighted masked concept supervision loss defined in Equations~(\ref{equ:concept_loss})--(\ref{equ:final_loss}), with $\lambda=1.0$.

\section{Results}
\begin{table*}[b]
\centering
\caption{Test-set performance for class~1 on the AFFECT dataset and the PHQ-9 depression dataset. For AFFECT, class~1 denotes high negative affect (NA); for PHQ-9, class~1 denotes follow-up-needed depressive state.}
\label{tab:performance_class1_datasets}
\begin{tabular}{llcccc}
\toprule
Dataset & Method & Precision & Recall & F1-score & AUC \\
\midrule
\multirow{4}{*}{AFFECT (NA)}
& \textbf{Proposed (CIT)} & 0.650 & \textbf{0.903} & \textbf{0.756} & \textbf{0.883} \\
& Transformer (no concepts) & \textbf{0.704} & 0.778 & 0.739 & 0.875 \\
& CIT (w/o concept loss) & 0.628 & 0.903 & 0.741 & 0.864 \\
& CIT + OCSVM Concept & 0.624 & 0.854 & 0.721 & 0.830 \\
\midrule
\multirow{4}{*}{PHQ-9 depression}
& \textbf{Proposed (CIT)} & 0.703 & \textbf{0.839} & \textbf{0.765} & 0.831 \\
& Transformer (no concepts) & \textbf{0.714} & 0.807 & 0.758 & 0.813 \\
& CIT (w/o concept loss) & 0.634 & 0.839 & 0.722 & 0.819 \\
& CIT + OCSVM Concept & 0.703 & 0.839 & 0.765 & \textbf{0.847} \\
\bottomrule
\end{tabular}
\end{table*}

\begin{figure}[t]
    \centering
    \includegraphics[width=\textwidth]{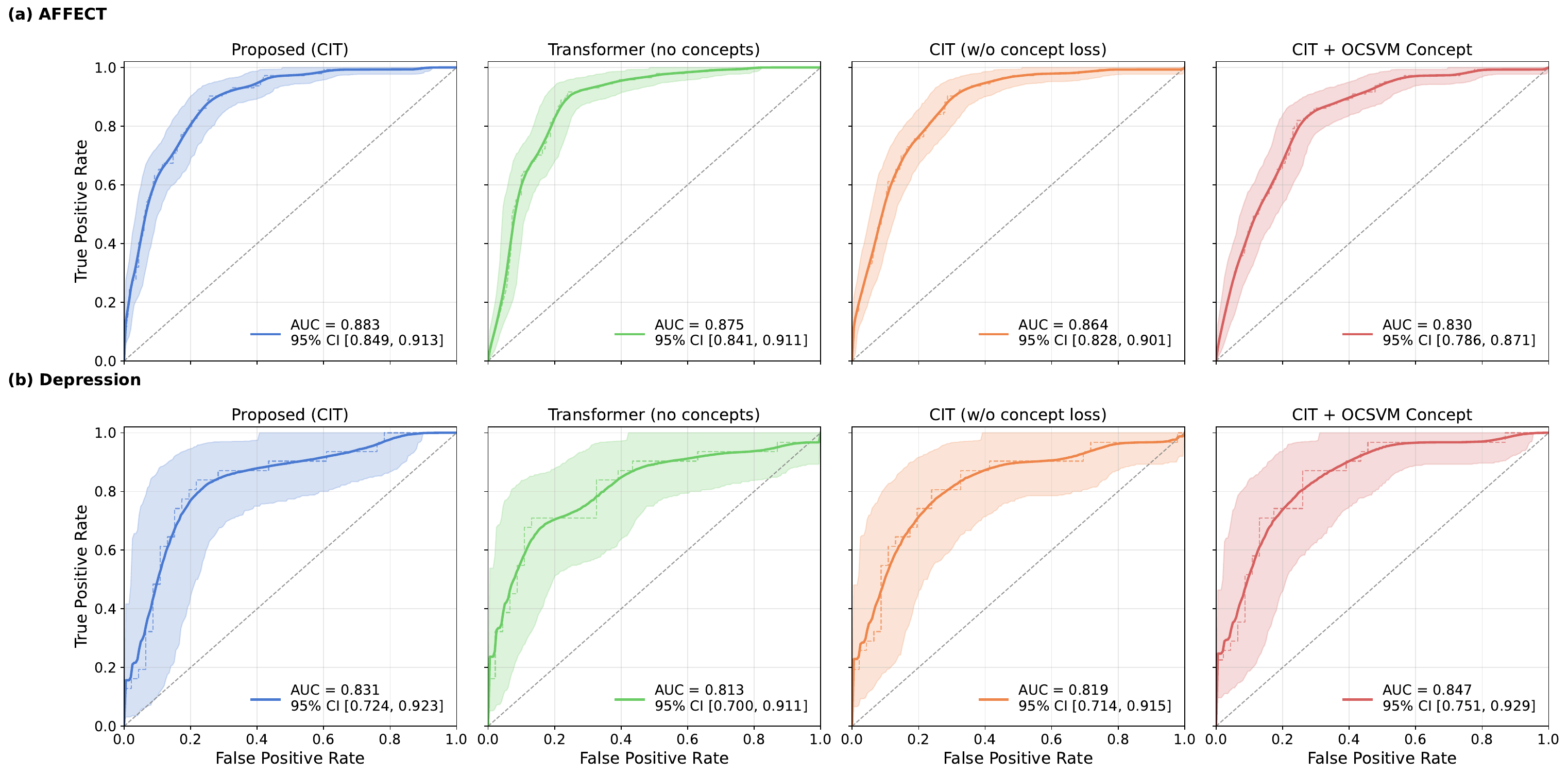}
    \caption{ROC analysis of the compared models on the two evaluation datasets. Panels show ROC curves for  (A) the AFFECT dataset and (B) the PHQ-9 depression dataset. Solid lines indicate ROC curves, and shaded areas indicate 95\% bootstrap confidence intervals. AUC values and 95\% confidence intervals are reported within each panel. }
    \label{fig:roc_curves}
\end{figure}
\subsection{Performance on the AFFECT Dataset}
 
On the AFFECT dataset, the proposed CIT achieved the highest F1-score (0.756), recall (0.903), and AUC (0.883) among the compared methods. Table~\ref{tab:performance_class1_datasets} summarizes the test-set performance for class~1, and Figure~\ref{fig:roc_curves} shows the corresponding ROC curves with 95\% confidence intervals.
Compared with the Transformer baseline without concepts, CIT improved class~1 recall (0.903 vs.~0.778), although its precision was lower (0.650 vs.~0.704). 
The resulting F1-score was higher for CIT than for the Transformer baseline (0.756 vs.~0.739). 
Removing the concept loss reduced the F1-score to 0.7407 and the AUC to 0.864. 
Replacing the LLM-derived concept targets with OCSVM-derived concept targets resulted in a lower F1-score (0.721) and AUC (0.830). 
The corresponding confusion matrices are provided in Supplementary Figure~S1.

\subsection{Performance on the PHQ-9 Depression Dataset}

We further evaluated the proposed framework on the PHQ-9 depression dataset. On this dataset, the proposed CIT achieved an F1-score of 0.765, with precision of 0.703, recall of 0.839, and AUC of 0.831. Table~\ref{tab:performance_class1_datasets} indicates the class~1 performance at the default decision threshold, and Figure~\ref{fig:roc_curves} demonstrates the ROC curves with 95\% confidence intervals. 
Compared with the Transformer baseline without concepts, CIT showed a slightly higher F1-score (0.765 vs.~0.758) and recall (0.839 vs.~0.807), while the Transformer baseline had slightly higher precision (0.714 vs.~0.703). 
CIT also achieved a higher F1-score than the model without concept loss (0.765 vs.~0.722). 
Compared with the OCSVM-based concept variant, CIT matched the fixed-threshold F1-score (0.765 vs.~0.765), precision (0.703 vs.~0.703), and recall (0.839 vs.~0.839), while the OCSVM-based variant achieved a slightly higher AUC (0.847 vs.~0.831). 
The corresponding confusion matrices are provided in Supplementary Figure~S2.

\subsection{Concept-level Explainability}

In this section, we perform the concept-level explainability analysis on the AFFECT dataset. 
Note that the PHQ-9 depression dataset was used as an experiment to assess applicability, and the corresponding concept-level visualizations are provided in Supplementary Figures~S3 and S4.

To clarify how the learned concept scores explain model behavior, we analyze them at three levels. 

\begin{figure*}[!htbp]
    \centering
    \includegraphics[width=\textwidth]{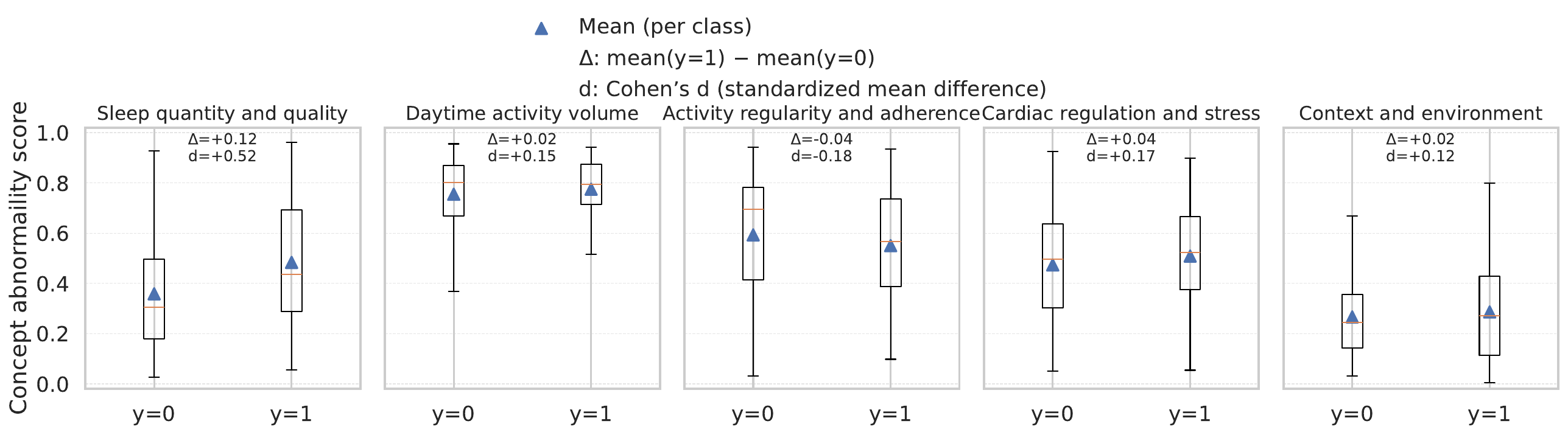}
    \caption{Class-level concept abnormality distributions on the AFFECT test set. Each panel shows the distribution of a learned concept abnormality score grouped by the ground-truth label. Blue triangles indicate per-class means. $\Delta$ denotes the class~1 minus class~0 mean difference, and $d$ denotes Cohen's $d$. }
    \label{fig:boxplot}
\end{figure*}

\paragraph{Class-level concept differences.} First, we assess class-level differences by comparing the distributions of concept abnormality scores between the two ground-truth classes.
Figure~\ref{fig:boxplot} shows the distribution of the window-level concept abnormality scores grouped by the ground-truth labels. 
We examine whether the learned concept scores separate test windows with high NA from those without.

Among the five concepts, sleep quantity and quality showed the clearest between-class separation, with higher abnormality scores in class~1 ($\Delta=+0.12$, Cohen's $d=+0.52$). 
Daytime activity volume and context and environment showed small positive shifts ($\Delta=+0.02$, $d=+0.15$ and $d=+0.12$, respectively). 
Cardiac regulation and stress also showed a small positive shift ($\Delta=+0.04$, $d=+0.17$), whereas activity regularity and adherence showed a slight negative shift ($\Delta=-0.04$, $d=-0.18$).

\begin{figure*}[!htbp]
    \centering
    \includegraphics[width=\textwidth]{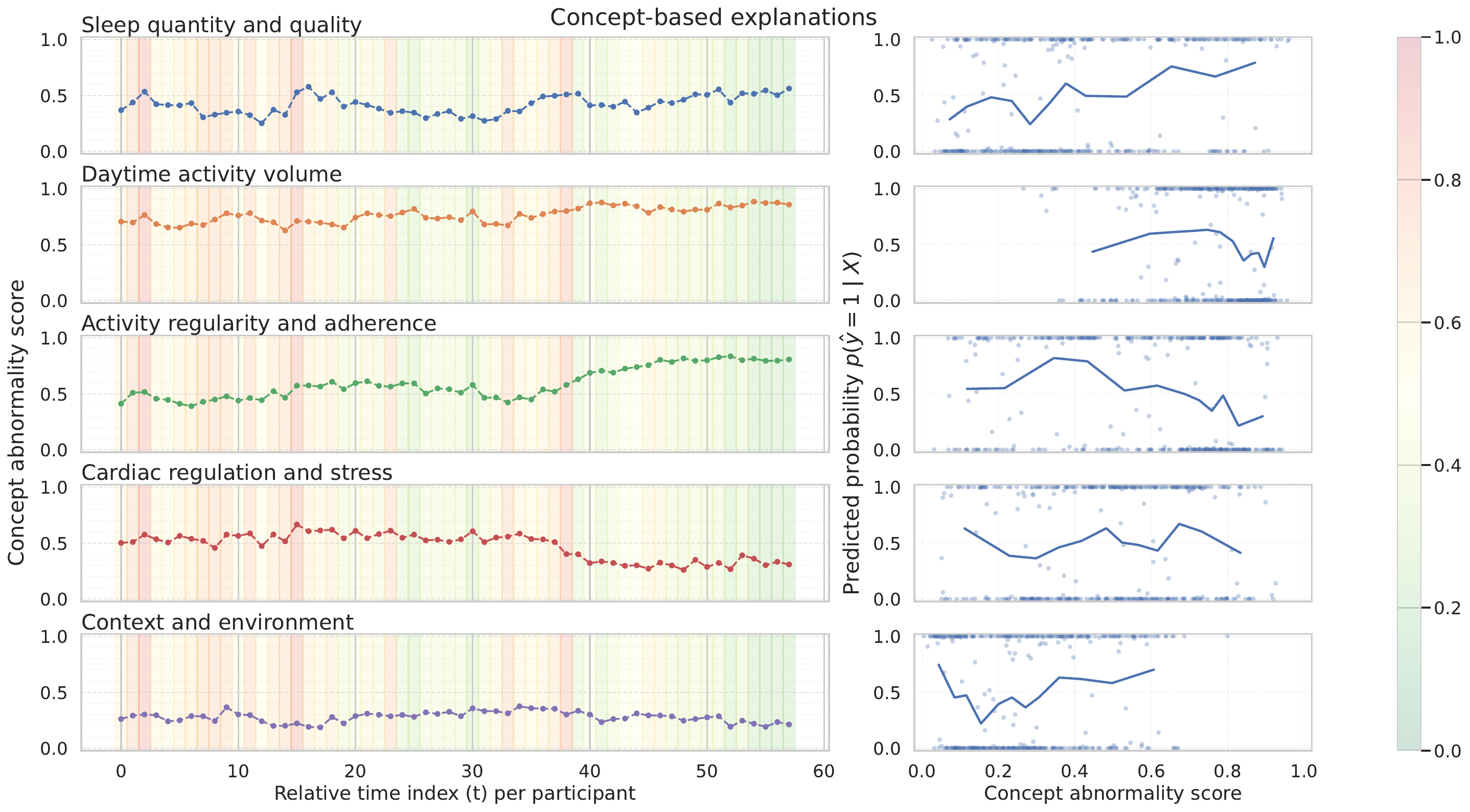}
    \caption{Population-level concept explanations on the AFFECT test set. The left column shows mean trajectories of the five learned concept abnormality scores over relative time.  Background shading indicates the predicted probability $p(\hat{y}=1\mid X)$, where green indicates $p(\hat{y}=1\mid X)$ closer to 0 and red indicates $p(\hat{y}=1\mid X)$ closer to 1. The right panels show the relationship between concept abnormality scores and predicted probability with a smoothed dependence curve.}
    \label{fig:trace}
\end{figure*}

\paragraph{Population-level concept--prediction relationships.} Second, we examine population-level model behavior by relating concept abnormality scores to the model-predicted probability of class~1 across the test set.
Figure~\ref{fig:trace} indicates a population-level view of how the learned concept abnormality scores relate to CIT's prediction outputs in the AFFECT test set. 
The left panels show mean concept abnormality trajectories over relative time, with background shading indicating the model-predicted probability of class~1. 
Greener regions correspond to lower predicted probability, whereas redder regions correspond to higher predicted probability. 
For example, activity regularity and adherence increase in the later part of the relative time axis, where the background is greener, suggesting that higher abnormality in this learned concept tends to appear during periods of lower predicted risk at the population level.
By placing the predicted probability behind the concept trajectories, the figure shows whether changes in each concept occur during periods of higher model-predicted risk. 
In this sense, the left panels help identify when a concept becomes elevated, whereas the right panels show whether higher values of that concept are associated with higher or lower model-predicted risk.

The right panels summarize the concept--prediction relationship more directly by plotting the predicted probability $p(\hat{y}=1\mid X)$ against each concept abnormality score with a smoothed dependence curve. 
Consistent with the temporal view, activity regularity and adherence show a negative relationship, with higher predicted risk concentrated at lower concept abnormality scores. 
In contrast, sleep quantity and sleep quality show the clearest positive relationship with the predicted probability of class~1, with higher predicted risk observed at higher concept abnormality scores. 
Daytime activity volume, cardiac regulation and stress, and context and environment show weaker or less monotonic relationships with the predicted probability.

\paragraph{Individual prediction example.} Third, we investigate an individual-level example to illustrate how an individual prediction can be summarized through the five learned concepts. 
In all analyses, higher concept abnormality scores indicate a stronger deviation in the corresponding behavioral or physiological concept as predicted by the concept head.
To illustrate how CIT summarizes an individual prediction, we examined one correctly classified positive test window from the AFFECT test set.
The example was selected as the correctly classified positive window whose logit was closest to the median logit among true-positive samples.

\begin{figure*}[!htbp]
    \centering
    \includegraphics[width=\textwidth]{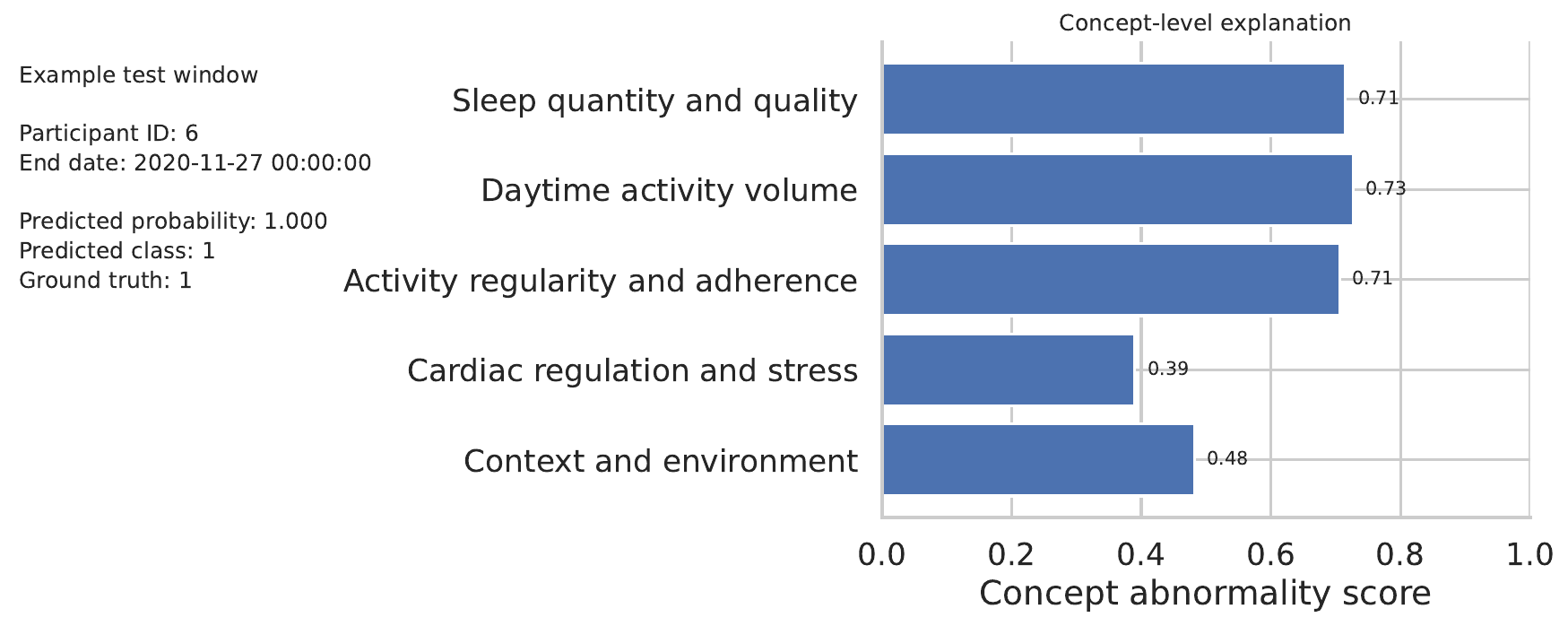}
    \caption{Individual-level concept explanation example from the AFFECT test set. The left panel summarizes the model output, including the predicted probability, predicted class, and ground-truth label. The right panel shows the five concept abnormality scores predicted by the concept head, where higher values indicate stronger abnormality in the corresponding behavioral or physiological concept. }
    \label{fig:participant_example}
\end{figure*}

As shown in Figure~\ref{fig:participant_example}, the selected window was from Participant~6 and ended on 27 November 2020.
CIT assigned a predicted probability of 1.000 to class~1, and the predicted class matched the ground-truth label.
The concept profile showed the highest abnormality score for daytime activity volume (0.73), followed by sleep quantity and quality (0.71) and activity regularity and adherence (0.71).
Context and environment showed a moderate abnormality score (0.48), while cardiac regulation and stress showed the lowest score (0.39).
Thus, for this positive prediction, the concept head summarized the window mainly through elevated activity-related and sleep-related abnormality scores.

\section{Discussion}

The results suggest that LLM-guided concept supervision can support class~1 detection in small-cohort longitudinal health-sensing settings, while also providing concept-level interpretability without requiring manual concept annotations. 
In the AFFECT experiment, CIT achieved the highest recall, F1-score, and AUC for detecting high negative affect among the compared methods. 
In the PHQ-9 depression experiment, CIT achieved the highest class~1 F1-score together with the OCSVM-based concept variant, while the OCSVM-based variant achieved a slightly higher AUC. 
Overall, the predictive benefit of LLM-derived concept supervision was clearest in the primary AFFECT experiment, while the PHQ-9 experiment showed comparable performance between the LLM-guided and OCSVM-based concept variants.

The form of improvement differed across the two datasets. 
In the AFFECT dataset, the performance gain was mainly reflected in improved sensitivity to class~1, with higher recall and F1-score than the Transformer baseline. 
This pattern is consistent with the design of CIT, where concept supervision is framed around deviations from participant-specific baseline patterns. 
In affective health monitoring, short-term deviations in sleep, activity, and related daily-life dynamics have been associated with affective functioning~\cite{Hickman2024SleepMoodAffect,Kalmbach2018DailyMood,jafarlou2023objective}. 
Thus, concept abnormality supervision may help the model attend to behavioral and physiological departures that are informative for high negative affect.

In contrast, the PHQ-9 experiment showed a more modest and less uniform improvement pattern. 
CIT slightly improved class~1 F1-score and recall over the Transformer baseline and the model without concept loss, but the OCSVM-based concept variant achieved a slightly higher AUC. 
This difference may reflect the nature of the target health outcome. 
Daily NA is a short-term affective state that may be closely coupled with acute changes in daily behavior, sleep, and physiological regulation~\cite{triantafillou2019relationship,do2024investigating}. 
PHQ-9-based categories, by contrast, summarize depressive-symptom severity over a longer recall period~\cite{kroenke2001phq}, and prior studies have linked depressive symptoms to behavioral and physiological patterns measured over broader monitoring windows~\cite{saeb2015mobile,zhang2021relationship}. 
As a result, the baseline-relative deviations used for LLM-guided concept abnormality scoring may align more directly with short-term affective-state prediction than with weekly depressive-symptom classification.

The comparison with the OCSVM-based concept variant provides a useful perspective on the role of the concept-supervision source. 
Although the LLM-derived concept targets outperformed the OCSVM-derived targets on the primary AFFECT dataset, the OCSVM-based variant achieved comparable class~1 performance and a slightly higher AUC on the PHQ-9 dataset. 
Given the limited size of the PHQ-9 dataset, this difference should be interpreted cautiously. 
Performance estimates, including AUC, can be imprecise in small-sample prediction studies~\cite{hajian2014sample}. 
Moreover, small training sets can increase the risk of variance and overfitting for more complex classifiers, whereas simpler or more strongly regularized learning signals may sometimes generalize more stably~\cite{raudys1991small}.

The concept-level analyses on the main AFFECT dataset clarify how the learned concept abnormality scores relate to both NA labels and model predictions. 
Across the class-level comparison and the population-level dependence plots, sleep quantity and quality showed the most consistent positive pattern. 
In the boxplot comparison, sleep-related abnormality was higher in class~1 windows, and in the dependence plots, higher sleep abnormality scores were associated with higher predicted probability of class~1. 
This suggests that sleep quantity and quality were the concept most strongly aligned with high negative affect in this dataset. 
This interpretation is consistent with prior affect monitoring literature, which has repeatedly identified sleep as a key correlate of affective functioning and day-to-day affective variation~\cite{Kalmbach2018DailyMood,Hickman2024SleepMoodAffect,jafarlou2023objective}.

Activity regularity and adherence showed a different but also consistent pattern. 
In the class-level comparison, this concept showed a slight negative shift, with higher abnormality scores observed in non-high negative affect windows. 
Similarly, the population-level dependence plot showed a negative relationship between this concept and the predicted probability of class~1. 
This pattern is plausible under our concept definition, where larger deviations in recent activity regularity may reflect greater behavioral variation and engagement rather than deterioration in affective state. 
Therefore, higher abnormality in this concept should not be interpreted as uniformly adverse. Its meaning depends on the behavioral content captured by the corresponding feature group.

The remaining concepts, including daytime activity volume, cardiac regulation and stress, and context and environment, showed weaker or less monotonic standalone relationships with the predicted probability. 
These weaker dependence patterns should not be interpreted as evidence that the concepts are irrelevant. 
In real-world sensing data, the affective meaning of activity, context, and physiological regulation may depend on interactions among multiple behavioral states, environmental conditions, and individual baselines~\cite{Puterman2017PhysicalActivityNA,Li2022ContextPhysicalActivity,Hachenberger2023HRVAffect,Gullett2023HRVReview}. 
Thus, these concepts may contribute to prediction in a conditional or interaction-dependent manner rather than through a simple monotonic association with predicted risk.

The individual prediction example serves a different purpose from the aggregate concept analyses. 
The class-level and population-level analyses examine how learned concept scores relate to labels and predicted risk across the test set, whereas the individual example examines whether the same concept space can be used to explain a specific prediction. 
This distinction is consistent with the motivation of local explanation methods such as LIME, where explanations are used to assess individual predictions rather than only global model behavior~\cite{ribeiro2016why}. 
However, CIT differs from post-hoc local surrogate methods in that the explanation space is learned as part of the model through concept supervision. 
Instead of explaining a prediction through feature contributions after training, CIT produces concept abnormality scores directly from the concept head. 
In this way, CIT connects aggregate concept analysis with prediction explanation: the same learned concept space can be used to both examine overall model behavior and to summarize individual model outputs.

Several limitations of this study should be noted. First, CIT relies on a predefined set of health-relevant concepts to organize LLM-guided concept supervision and test-time interpretation. The five concepts used in this study were designed to cover common behavioral, physiological, and contextual domains in wearable and mobile health sensing, but they represent one possible concept taxonomy rather than a universally optimal set. Alternative concept definitions may be more appropriate for different health outcomes or sensing modalities. Second, although CIT is designed for small-cohort longitudinal sensing studies, its generalizability should be further evaluated in more diverse cohorts, including populations with different demographic characteristics, sensing devices, data collection protocols, and health-related outcomes.

Future work should evaluate the proposed framework across additional datasets and health-related prediction tasks, and further validate the learned concept scores against domain knowledge or expert assessment. It would also be valuable to explore alternative concept definitions, stronger supervision strategies, and richer forms of multimodal grounding for concept learning.

\section{Conclusions}

In this paper, we proposed CIT, a Concept-Integrated Transformer framework with annotation-free LLM guidance for explainable prediction from mobile health-sensing data.
By transforming baseline-aware statistical summaries into concept-level abnormality supervision, CIT incorporates LLM-informed concept learning into a Transformer-based prediction framework without requiring manual concept annotations. Evaluations on the AFFECT dataset and a PHQ-9 depression dataset showed that CIT achieved competitive predictive performance while producing structured concept-level explanations. 
On the AFFECT dataset, CIT achieved the highest recall, F1-score, and AUC for detecting high negative affect among the compared methods. 
On the PHQ-9 dataset, CIT achieved the highest class~1 F1-score together with the OCSVM-based concept variant, while the OCSVM-based variant achieved a slightly higher AUC. 
The concept-level analyses further showed that the learned concept abnormality scores could summarize model behavior in terms of interpretable behavioral and physiological dimensions. 
In the AFFECT experiment, sleep quantity and quality emerged as the clearest population-level signal, and the learned concept scores showed distinct relationships with model predictions. 
Overall, these findings suggest the potential of LLM-guided concept supervision for developing explainable predictive models in small-cohort longitudinal health-sensing settings.

\section*{Code availability}

The source code used to implement the proposed method will be made publicly available upon publication.

\bibliography{sample}

\section*{Author contributions statement}


Y.W. conceived the study, designed the method, conducted the experiments, analysed the results, and drafted the manuscript. I.A., A.M.R., and P.L. supervised the study and provided feedback on the method, experiments, and manuscript. All authors reviewed and approved the final manuscript.

\section*{Funding}
The authors declare that no specific funding was received for this work.

\section*{Ethics declaration}
This study used publicly available, de-identified datasets and did not involve new recruitment of human participants or collection of identifiable personal data. Ethical approval and informed consent for the original data collection were obtained by the investigators of the respective source studies, as described in the corresponding dataset publications. No additional ethical approval was required for the secondary analyses conducted in this study.

\clearpage

\lstset{
  basicstyle=\ttfamily\small,
  breaklines=true,
  breakatwhitespace=false,
  columns=fullflexible,
  keepspaces=true,
  frame=single
}
\begin{center}
{\Large \textbf{Supplementary Information}}\\
\end{center}

\renewcommand{\thefigure}{S\arabic{figure}}
\renewcommand{\thetable}{S\arabic{table}}
\setcounter{figure}{0}
\setcounter{table}{0}

\section*{Confusion Matrix Visualizations}
\begin{figure*}[h]
    \centering
    \includegraphics[width=\textwidth]{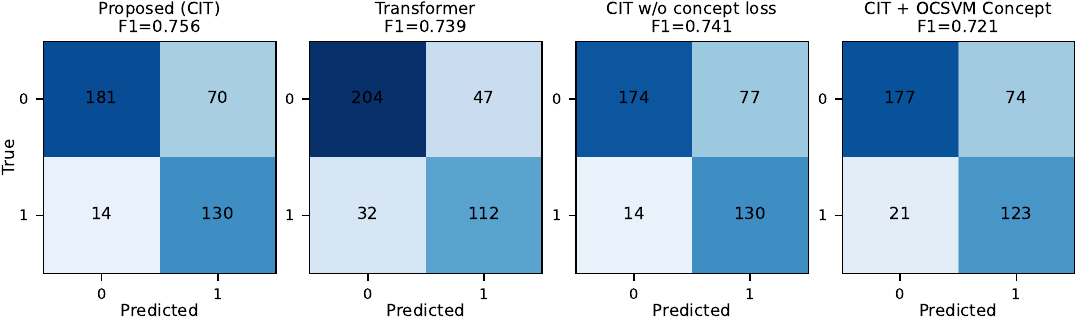}
    \caption{Test-set confusion matrices on the AFFECT dataset for the proposed CIT, the Transformer baseline, CIT without concept loss, and CIT with OCSVM-based concept targets.}
    \label{fig:CM}
\end{figure*}

\begin{figure*}[h]
    \centering
    \includegraphics[width=\textwidth]{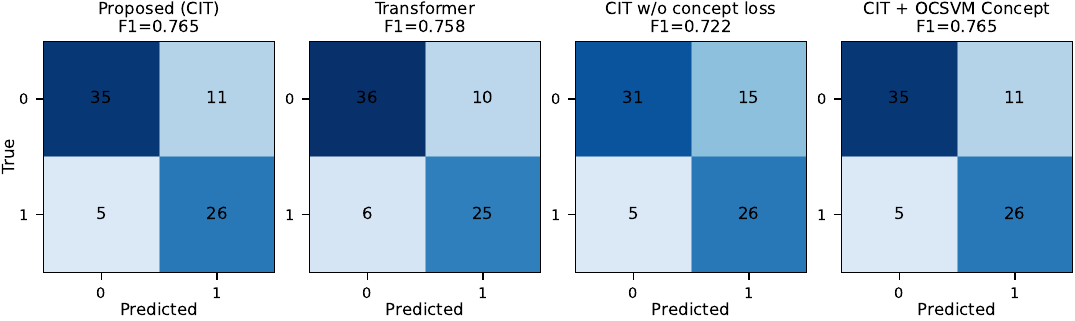}
    \caption{Test-set confusion matrices on the PHQ-9 depression dataset for the proposed CIT, the Transformer baseline, CIT without concept loss, and CIT with OCSVM-based concept targets.}
    \label{fig:cm_phq9}
\end{figure*}

Figure~\ref{fig:CM} shows the confusion matrix of AFFECT dataset, and Figure~\ref{fig:cm_phq9} shows the confusion matrix of PHQ-9 depression dataset.

\section*{Supplementary Concept-level Visualizations for the PHQ-9 Experiment}
\label{sec:supp_phq9_concept_visualization}

\begin{figure}[h]
    \centering
    \includegraphics[width=\textwidth]{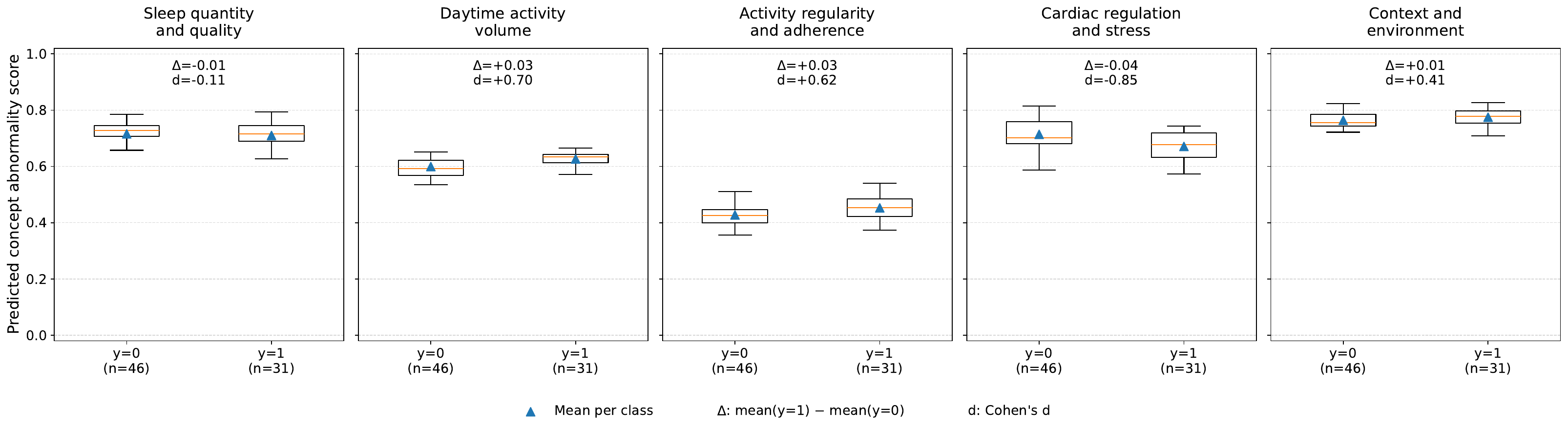}
    \caption{\textbf{Predicted concept abnormality scores on the held-out PHQ-9 test set.}
    The scores are stratified by the binary PHQ-9-derived label. Class~0 denotes "None-minimal", and class~1 denotes "follow-up needed". }
    \label{fig:supp_phq9_boxplot}
\end{figure}

\begin{figure}[h]
    \centering
    \includegraphics[width=\textwidth]{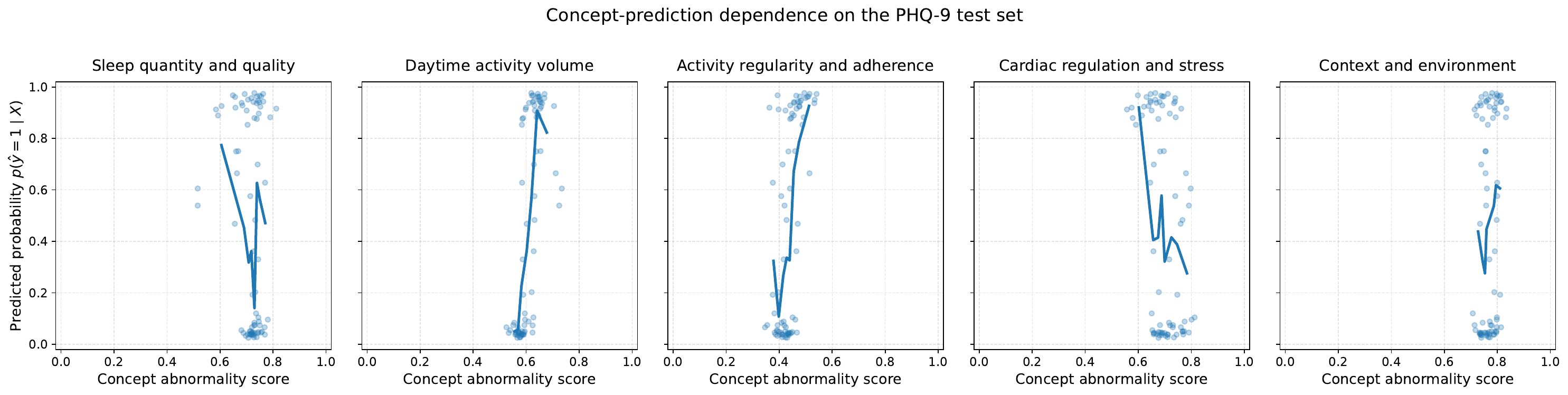}
    \caption{\textbf{Concept--prediction dependence plots on the held-out PHQ-9 test set.}
    Each panel shows the relationship between a predicted concept abnormality score and the model-predicted probability of class~1. The smoothed curve summarizes the average predicted probability across concept-score bins.}
    \label{fig:supp_phq9_dependence}
\end{figure}

Figure~\ref{fig:supp_phq9_boxplot} shows the distribution of predicted concept abnormality scores on the held-out PHQ-9 test set, stratified by the binary PHQ-9-derived label. Figure~\ref{fig:supp_phq9_dependence} shows the dependence between the predicted concept scores and the predicted probability of class~1. The concept scores were inferred from the trained CIT concept head without LLM calls at test time.

\clearpage

\section*{Feature-to-Concept Mapping}
\label{sec:supp_feature_concept}

Table~\ref{tab:feature_concept_mapping} summarizes how the 43 daily features in AFFECT dataset were assigned to the five predefined concepts used in CIT. Table~\ref{tab:depression_feature_concept_mapping} summarizes the corresponding mapping for the PHQ-9 depression dataset.

\begin{longtable}{p{0.26\textwidth}p{0.68\textwidth}}
\caption{Feature-to-concept mapping used for concept construction (AFFECT dataset).}
\label{tab:feature_concept_mapping}\\
\toprule
\textbf{Concept} & \textbf{Feature description} \\
\midrule
\endfirsthead

\toprule
\textbf{Concept} & \textbf{Feature description} \\
\midrule
\endhead

\bottomrule
\endfoot

Sleep quantity and quality &
\begin{tabular}[t]{@{}l@{}}
Awake time during the sleep period \\
REM sleep duration \\
Light sleep duration \\
Deep sleep duration \\
Total sleep duration
\end{tabular} \\
\midrule

Daytime activity volume &
\begin{tabular}[t]{@{}l@{}}
Daily movement volume \\
Low-intensity activity duration \\
Medium-intensity activity duration \\
High-intensity activity duration \\
MET expenditure during inactive periods \\
MET expenditure during low-intensity activity \\
MET expenditure during medium-intensity activity \\
MET expenditure during high-intensity activity \\
Activity-related energy expenditure \\
Total daily energy expenditure \\
Traveled distance \\
Number of running steps \\
Number of walking steps \\
Total step count \\
Average MET level
\end{tabular} \\
\midrule

Activity regularity and adherence &
\begin{tabular}[t]{@{}l@{}}
Stay active score \\
Movement every hour score \\
Training volume score \\
Daily target attainment score \\
Training frequency score \\
Recovery time score \\
Inactivity alerts
\end{tabular} \\
\midrule

Cardiac regulation and stress &
\begin{tabular}[t]{@{}l@{}}
Average heart rate \\
Average RMSSD \\
Standard deviation of heart rate \\
Standard deviation of RMSSD
\end{tabular} \\
\midrule

Context and environment &
\begin{tabular}[t]{@{}l@{}}
Minimum atmospheric pressure \\
Maximum atmospheric pressure \\
Location change \\
Activity level \\
Number of activity names \\
Number of activity types \\
Most frequent activity type \\
Most frequent activity name \\
Weekday \\
Month \\
Day \\
Hour
\end{tabular} \\
\end{longtable}

\clearpage

\begin{longtable}{p{0.26\textwidth}p{0.68\textwidth}}
\caption{Feature-to-concept mapping used for concept construction (depression dataset).}
\label{tab:depression_feature_concept_mapping}\\
\toprule
\textbf{Concept} & \textbf{Feature description} \\
\midrule
\endfirsthead

\toprule
\textbf{Concept} & \textbf{Feature description} \\
\midrule
\endhead

\bottomrule
\endfoot

Sleep quantity and quality &
\begin{tabular}[t]{@{}l@{}}
Mean sleep heart rate \\
Mean sleep RMSSD \\
SD sleep heart rate \\
SD sleep RMSSD \\
Slope sleep heart rate \\
Slope sleep RMSSD \\
Intercept sleep heart rate \\
Intercept sleep RMSSD \\
Awake time during sleep \\
Average respiratory rate \\
Sleep efficiency \\
Average sleep heart rate \\
Lowest sleep heart rate \\
Latency of falling asleep \\
Amount of REM sleep \\
Average sleep RMSSD \\
Sleep skin temperature deviation from long-term average \\
Sleep skin temperature deviation from environment \\
Total sleep time
\end{tabular} \\
\midrule

Daytime activity volume &
\begin{tabular}[t]{@{}l@{}}
Mean activity level \\
Average MET \\
Total calories \\
Minutes with high intensity activity \\
Minutes being inactive \\
Minutes with low intensity activity \\
Minutes with medium intensity activity \\
Minutes user not wearing the ring \\
Minutes spent resting \\
Number of steps
\end{tabular} \\
\midrule

Activity regularity and adherence &
\begin{tabular}[t]{@{}l@{}}
SD activity level \\
Slope activity level \\
Intercept activity level \\
Number of places \\
Home duration \\
Outdoor duration \\
Mean of outdoor duration \\
Standard deviation of outdoor duration \\
Total travel distance
\end{tabular} \\
\midrule

Cardiac regulation and stress &
\begin{tabular}[t]{@{}l@{}}
HR \\
HRV MeanNN \\
HRV SDNN \\
HRV SDANN1 \\
HRV SDNNI1 \\
HRV RMSSD \\
HRV SDSD \\
HRV CVNN \\
HRV CVSD \\
HRV MedianNN \\
HRV MadNN \\
HRV MCVNN \\
HRV IQRNN \\
HRV pNN50 \\
HRV pNN20 \\
HRV HTI \\
HRV TINN \\
HRV LF \\
HRV HF \\
HRV VHF \\
HRV LFHF \\
HRV LFn \\
HRV HFn \\
HRV LnHF \\
HRV SD1 \\
HRV SD2 \\
HRV SD1SD2 \\
HRV S
\end{tabular} \\
\midrule

Context and environment &
\begin{tabular}[t]{@{}l@{}}
Income call durations \\
Outgoing call duration \\
Missed call duration \\
Voicemail call duration \\
Income call counts \\
Outgoing call counts \\
Missed call counts \\
Voicemail counts \\
Number of received messages \\
Number of sent messages \\
Number of notifications from applications type: Productivity \\
Number of notifications from applications type: Photography \\
Number of notifications from applications type: Communication \\
Number of notifications from applications type: Lifestyle \\
Number of notifications from applications type: Auto \& Vehicles \\
Number of notifications from applications type: Travel \& Local \\
Number of notifications from applications type: Education \\
Number of notifications from applications type: Finance \\
Number of notifications from applications type: Video Players \& Editors \\
Number of notifications from applications type: Social \\
Number of notifications from applications type: Books \& Reference \\
Number of notifications from applications type: Shopping \\
Number of notifications from applications type: Health \& Fitness \\
Number of notifications from applications type: Entertainment \\
Number of notifications from applications type: Business \\
Number of notifications from applications type: Music \& Audio \\
Number of notifications from applications type: unknown \\
Number of notifications from applications type: Tools \\
Number of screen off \\
Number of screen on \\
Number of screen locks \\
Number of screen unlocks \\
Mean of battery charge \\
Number of battery charger plugins \\
Variance of latitude \\
Variance of speed \\
Mean of speed
\end{tabular} \\

\end{longtable}

\clearpage

\section*{Original Prompt Used for LLM-Guided Concept Abnormality Supervision}
\label{sec:supp_prompt}

This section provides the original prompt used to generate LLM-guided concept abnormality supervision targets. During implementation, the input evidence section of the prompt was populated with automatically generated summary statistics for each 7-day window and its corresponding 28-day participant-specific baseline. The ground-truth negative affect label was not provided to the LLM.

\subsection*{Original LLM Prompt for the AFFECT Dataset}

\begin{lstlisting}
SYSTEM_PROMPT = """You are an expert in digital phenotyping and behavioral signal analysis.

You will be given a summary table for ONE participant over a 7-day window.
The table includes window stats (mean/std/slope/last/missing) and, when available,
baseline-relative stats (z-scores and percent-changes) relative to this participant's own historical baseline.

Rules:
1) Base your judgments ONLY on the provided summary table. Do NOT use unstated commonsense assumptions.
2) Do NOT infer or mention any affect/label/outcome. This task is label-free.
3) If evidence is insufficient/ambiguous (e.g., many nulls, high missing, contradictory signals, no usable baseline),
   you MUST output null abnormality for that concept.
4) evidence_keys MUST be keys that appear verbatim in the summary table.
"""

USER_TEMPLATE = """Summary table (each line is "key: value  # description"):

{summary_text}

High-level behavioral concepts (abnormality = deviation from this participant's baseline/typical pattern):
- sleep_quantity_quality: sleep duration/architecture (total/awake/REM/light/deep), stability and acute shifts.
- daytime_activity_volume: activity amount/intensity (steps, distance, MET, calories, minutes in intensity bands).
- activity_regular_adherence: adherence/regularity (scores, move-every-hour, inactivity alerts, goal progress).
- cardio_regulation_stress: HR/HRV regulation and stability (heart_rate, RMSSD, hr_std, rmssd_std).
- context_environment: contextual background (time/calendar, activity codes, location changes, pressure/weather).

Important constraint for context_environment:
- Calendar/code variables (weekday/month/day/hour and activity-code fields) are BACKGROUND ONLY.
  Prefer physiological/behavioral evidence when available.

Task:
For this sample, estimate the ABNORMALITY of each concept within this 7-day window.

Return ONE JSON object with EXACTLY these keys and schema:

{
  "sleep_quantity_quality": {"abnormal": 0.0, "confidence": 0.0, "evidence_keys": []},
  "daytime_activity_volume": {"abnormal": 0.0, "confidence": 0.0, "evidence_keys": []},
  "activity_regular_adherence": {"abnormal": 0.0, "confidence": 0.0, "evidence_keys": []},
  "cardio_regulation_stress": {"abnormal": 0.0, "confidence": 0.0, "evidence_keys": []},
  "context_environment": {"abnormal": 0.0, "confidence": 0.0, "evidence_keys": []}
}

Where:
- abnormal is a float in [0,1] OR null.
- abnormal=0 means no meaningful deviation; abnormal=1 means extreme deviation.
- confidence is a float in [0,1].
- If abnormal is NOT null: evidence_keys MUST contain 2-4 keys from the table.
- If abnormal is null: evidence_keys MUST be [] and confidence MUST be <= 0.3.

Output valid JSON only. No extra text.
"""
\end{lstlisting}

\clearpage

\subsection*{Original LLM Prompt for the Depression Dataset}
\label{sec:supp_depression_prompt}

\begin{lstlisting}
SYSTEM_PROMPT_V3 = """You are an expert in digital phenotyping and behavioral signal analysis.

You will be given a summary table for ONE participant over a 7-day window before a weekly questionnaire assessment.
The table includes current-window statistics and, when available, baseline-relative statistics relative to this participant's own historical baseline.

Rules:
1) Base your judgments ONLY on the provided summary table. Do NOT use unstated commonsense assumptions.
2) Do NOT infer or mention depression, PHQ-9, affect, diagnosis, label, or outcome. This task is label-free.
3) Judge abnormality as deviation from the participant's own recent baseline or, when baseline is unavailable, as unusual within-window variability/trend/missingness based only on the table.
4) If evidence is insufficient or ambiguous, including many nulls, high missingness, contradictory signals, or no usable evidence for a concept, output null abnormality for that concept.
5) evidence_keys MUST be keys that appear verbatim in the summary table.
6) Do not use feature names alone as evidence. Use derived keys such as zBase_..., zBaseMean7_..., pctBase_..., pctBaseLast_..., mean7_..., std7_..., slope7_..., last_..., or miss7_....
"""


USER_TEMPLATE_V3 = """Summary table (each line is "key: value  # description"):

{summary_text}

High-level behavioral concepts. Abnormality means deviation from this participant's own baseline or typical recent pattern:

- sleep_quantity_quality:
  Sleep quantity and quality, including total sleep time, awake time during sleep, sleep efficiency, sleep latency, REM sleep, sleep respiratory rate, sleep heart rate, sleep RMSSD, and sleep temperature deviations.

- daytime_activity_volume:
  Overall daytime activity amount and intensity, including activity level, MET, calories, steps, inactive time, resting time, low/medium/high intensity activity minutes, and non-wear time.

- activity_regular_adherence:
  Activity and mobility regularity, including variability or trend in activity level, home/outdoor duration, number of places, travel distance, and stability of mobility routines.

- cardio_regulation_stress:
  Cardiovascular and autonomic regulation, including HR, HRV time-domain features, HRV frequency-domain features, LF/HF balance, RMSSD, SDNN, pNN50/pNN20, and related HRV geometry features.

- context_environment:
  Contextual and digital behavioral environment, including calls, messages, notifications, screen interaction, battery/charging behavior, speed, latitude variance, and other smartphone-derived context signals.

Important constraint for context_environment:
- Smartphone and mobility features can be used as evidence when they appear in the table.
- Do not treat mere availability of phone data as abnormal. Use only deviations, trends, variability, or strong current-window patterns shown by the summary keys.

Task:
For this sample, estimate the ABNORMALITY of each concept within this 7-day window.

Return ONE JSON object with EXACTLY these keys and schema:

{
  "sleep_quantity_quality": {"abnormal": 0.0, "confidence": 0.0, "evidence_keys": []},
  "daytime_activity_volume": {"abnormal": 0.0, "confidence": 0.0, "evidence_keys": []},
  "activity_regular_adherence": {"abnormal": 0.0, "confidence": 0.0, "evidence_keys": []},
  "cardio_regulation_stress": {"abnormal": 0.0, "confidence": 0.0, "evidence_keys": []},
  "context_environment": {"abnormal": 0.0, "confidence": 0.0, "evidence_keys": []}
}

Where:
- abnormal is a float in [0,1] OR null.
- abnormal=0 means no meaningful deviation; abnormal=1 means extreme deviation.
- confidence is a float in [0,1].
- If abnormal is NOT null: evidence_keys MUST contain 2-4 keys from the table.
- If abnormal is null: evidence_keys MUST be [] and confidence MUST be <= 0.3.
- Prefer baseline-relative keys when available, especially zBase_..., zBaseMean7_..., pctBase_..., and pctBaseLast_....
- If baseline-relative keys are null, you may use mean7_..., std7_..., slope7_..., last_..., and miss7_... only when they provide sufficient evidence.

Output valid JSON only. No extra text.
"""
\end{lstlisting}


\end{document}